\pdfoutput=1
\documentclass[11pt]{article}

\usepackage[final]{acl}

\usepackage{times}
\usepackage{latexsym}

\usepackage[T1]{fontenc}
\usepackage[utf8]{inputenc}

\usepackage{microtype}

\usepackage{inconsolata}

\usepackage{fancyvrb}
\usepackage{graphicx}
\usepackage{tikz}
\usepackage{enumerate}
\usepackage{pifont}
\newcommand{\cmark}{\ding{51}}%
\newcommand{\xmark}{\ding{55}}
\usepackage{algorithm}
\usepackage{algpseudocode}
\usepackage{makecell}
\usepackage{soul}
\usepackage{listings}
\usepackage{courier}
\usepackage{algorithm}
\usepackage{algorithmicx}

\usepackage{todonotes}

\definecolor{gold}{rgb}{1, 0.65, 0.0}
\definecolor{bpurple}{rgb}{0.6, 0.0, 1.0}

\title{The Effect of Data Partitioning Strategy on Model Generalizability: A Case Study of Morphological Segmentation}

\author{Zoey Liu \\
  Department of Linguistics \\
  University of Florida \\
  \texttt{liu.ying@ufl.edu} \\\And
  Bonnie J. Dorr \\
  Computer \& Information Science \& Engineering \\
  Florida Institute for National Security\\
  University of Florida \\
  \texttt{bonniejdorr@ufl.edu} \\}

\begin{document}
\maketitle
\begin{abstract}
Recent work to enhance data partitioning strategies for more realistic model evaluation 
%in order to probe for generalizability %ability to probe for generalizability 
face challenges in providing a clear optimal choice. This study addresses these challenges, focusing on morphological segmentation and synthesizing limitations related to language diversity, adoption of multiple datasets and splits, and detailed model comparisons. Our study leverages data from 19 languages, including ten indigenous or endangered languages across 10 language families with diverse morphological systems (polysynthetic, fusional, and agglutinative) and different degrees of data availability. 
%\bonnieshort{What does "when facing new data" mean?  I'm actually having trouble distinguishing (1) and (2) as they both involve "new data".  Maybe this could be revised as follows? "Our results show that, when faced with new data: (1) random splits yield models with better performance; (2) model rankings derived from random splits tend to generalize more consistently." (I left commented out the original in case you want to bring it back.)}
We conduct large-scale experimentation with varying sized combinations of training and evaluation sets as well as new test data. 
%Our results show that: (1) random splits yield models with better performance when facing new data; (2) model rankings derived from random splits tend to generalize more consistently to new data. 
%% Editing above:
Our results show that, when faced with new test data: 
%(1) random splits yield models with better performance; 
(1) models trained from random splits are able to achieve higher numerical scores;
(2) model rankings derived from random splits tend to generalize more consistently.
%; and (3) dataset-specific features are predictive of model performance, when controlling for the specific data partitioning strategy.

\end{abstract}

\section{Introduction}

Evaluations of computational models in natural language processing typically rely on a single dataset, though there are exceptions, such as high-resource languages like English, which have multiple datasets available for specific tasks~\cite{rajpurkar-etal-2016-squad,kwiatkowski-etal-2019-natural,warstadt-etal-2020-blimp-benchmark}. Such datasets are typically provided with \textit{one} partitioning, including at least a training and test set, with an optional validation set depending on data availability~\cite{gauthier-etal-2016-collecting,cotterell-etal-2015-labeled}. The single data split is typically determined by shared task organizers~\cite{kurimo2006unsupervised}, benchmark designers~\cite{wang-etal-2018-glue}, and ground-breaking paper authors~\cite{collins-2002-discriminative}. The rationale behind the particular data partition, however, is not always clear or mentioned at all (cf. \citet{de-marneffe-etal-2021-universal}).

Recent work has called into question the adoption of one data split~\cite{gorman-bedrick-2019-need,liu-etal-2023-investigating,kodner-etal-2023-morphological} or  one dataset~\cite{sogaard-etal-2021-need} for model evaluation.
These previous studies point out that individual model performance as well as model rankings derived from just one single split of training-(validation)-test sets \textit{may fail to generalize} when applied to an alternative split of the same dataset, or even to new unseen data from the same domain. 
Thus, drawing conclusions based on a single data partition has the potential for being unreliable.  

This study investigates the effect of different data partitioning methods on model generalizability in cross-linguistic scenarios with \textit{varying data availability}. We use morphological segmentation as the testbed, i.e., the task of decomposing a word into its component morphemes (\textit{avocados} $\rightarrow$ \textit{avocado} + \textit{s}). While a range of studies have been undertaken to explore the impact of data partitioning strategies on generalizability of model performance (see Section~\ref{sec:related-work} for some examples), 
it is safe to say that no consensus has been reached regarding the specific choices of data partition strategies. This is largely due to the fact that these studies face several important limitations, which we describe below.

\noindent \textbf{Lack of language diversity} First, a considerable portion of prior work \cite{gorman-bedrick-2019-need,sogaard-etal-2021-need} predominantly focuses on English (cf. \citet{bender2019benderrule,DBLP:conf/lrec/DucelFLL22}). It is possible, however, that  an optimal data partitioning strategy, if one exists, is dependent on the languages (and tasks) under investigation. This is because the typological traits of the languages can have an impact on the distributions of the resulting training/test sets and new unseen data. For example, if languages exhibit greater morphological regularity, alternative data partitioning approaches might yield comparable model performance.

\noindent \textbf{Lack of multiple datasets and data splits}  Building on the first point, the tasks investigated in previous literature often enjoy ample data availability. For high-resource languages with abundant data for a given task, it is often assumed (implicitly or explicitly) that the  selected dataset or data split adequately represents the task or language.
Therefore, a data partitioning strategy that fares well on the same dataset or data split is expected to yield models that generalize reasonably to new unseen data, particularly from the same domain. 
However, in scenarios with constraints on data availability, the representativeness of the chosen dataset or data split becomes questionable. 
What kind of data partitioning strategy is appropriate to apply when facing different extents of data availability thus remains an open question.
%The appropriate data partitioning strategy to apply in resource-restrained settings thus remains an open question. 
\citet{liu-etal-2023-investigating} address the aforementioned issues to some extent, but lack evaluation of
trained models on new test samples. %, focusing instead on evaluating models using different training-evaluation splits.
Although \citeauthor{sogaard-etal-2021-need}\ recommend inclusion of multiple test sets, they do not consistently experiment accordingly for each task.

\noindent \textbf{Lack of model comparisons} Finally, while \citeauthor{gorman-bedrick-2019-need}\ compare a number of POS taggers, \citeauthor{sogaard-etal-2021-need}\  and \citeauthor{liu-etal-2023-investigating}\ apply one model for each task. Thus, they fail to provide a detailed analysis of model rankings and how these rankings may be affected by different partitioning strategies. It remains unclear if these rankings would still hold when considering new test samples.

Taking a data-driven approach, this work transcends that of prior approaches in several respects: 
\hspace*{-.1in}
\begin{tabular}{lp{2.75in}}
    $\bullet$&We attend to a typologically diverse set of 19 languages from ten language families, covering polysynthetic, fusional, and agglutinative morphological systems. These languages have different amounts of data available pertaining  to morphological segmentation. In addition, ten of these languages are indigenous or endangered languages, painting a typologically rich set of language samples for our study. \\
    $\bullet$&We compare four model architectures in order to analyze model rankings. \\
    $\bullet$&Perhaps most importantly, we conduct a sequence of large-scale experiments, varying both the combinations of training and evaluation sets along with their respective sizes. To evaluate the generalizations of model performance resulting from different data partitioning strategies, we generate new test samples of different sizes as well. 
\end{tabular}

\noindent In what follows, Section~\ref{sec:related-work} describes recent studies to explore the respective effect of different data splits on model performance. Section~\ref{sec:experiments} presents our experimentation, 
%with 
including %the creation of
dataset creation
%creation, as well as 
and evaluation of
four model architectures %that we assess 
to probe model generalizability. Section~\ref{sec:analysis} provides an analysis, answering questions about the impact of data partitioning strategies on model generalizability. Section~\ref{sec:discussion} concludes with possible 
avenues for future work. Finally, we address the limitations
%weaknesses 
of our approach,
%in Section~\ref{sec:limitations}, which 
followed by a statement on ethics and %ethics statement on 
broader impact.

\section{Related Work}
\label{sec:related-work}

%\zoey{decided to swap out ``standard" for ``pre-defined"}

\noindent \textbf{Data partitioning strategies} Recent %work 
research proposes
%has proposed 
different strategies to address the question of data split impact on model generalizability (see Table~\ref{priorwork} for a summary of comparisons between previous work and our studies.). 

\begin{table}[h!]
%\footnotesize
\resizebox{\columnwidth}{!}{
%\centering
\begin{tabular}{lllll}
\hline
\textbf{Study} & \textbf{Multilingual} & \textbf{Including Resource-} & \textbf{Multi-} & \textbf{Multi-}  \\
 & & \textbf{constrained scenarios} & \textbf{datasets} & \textbf{models}   \\
\hline
G\&B & \xmark & \xmark & \xmark & \cmark \\
SEBF & \xmark & \xmark & not always  & \xmark  \\
& & & & \\
LSP & \cmark & \cmark & \xmark & \xmark   \\
Ours & \cmark & \cmark & \cmark & \cmark  \\
\hline
\end{tabular}}
\caption{\label{priorwork}
Comparisons of experimental setups between G\&B~\citep{gorman-bedrick-2019-need}, SEBF~\citep{sogaard-etal-2021-need}, LSP~\citep{liu-etal-2023-investigating}, and our study here. 
%\textcolor{blue}{Zoey: there has been some recent work including my own showing that there can be large score variability across seeds; if time permitting, I will add more seeds for more analysis (3 seeds for now); if were to do this, this will be another plus that previous work does not have}
}
\end{table}

\citet{gorman-bedrick-2019-need} conduct a series of replication and reproduction experiments on part-of-speech (POS) tagging using the Wall Street Journal (WSJ) from the Penn Treebank~\citep{marcus-etal-1993-building}.
Their work re-evaluates the performance of eight POS taggers previously claimed to achieve state-of-the-art performance on one split of the WSJ dataset from \citet{collins-2002-discriminative}. They refer to this as ``standard split'', dividing the WSJ dataset as follows: 00–18 as the training set, 19-21 as the development set, and 22-24 for test.
%For example, in the work of \citet{collins-2002-discriminative}, the training set consists of the following WSJ divisions: 00–18 (75\%) for training, 19-21 (12.5\%) for development, and 22-24 (12.5\%) for test. This split falls squarely within the range for ``standard split'' (which has traditionally been 60-80\%  for training, 10-20\% for validation, and 10-20\% for the test set, and until recently went unchallenged.
%The exploration of \citet{gorman-bedrick-2019-need} compares
%They explore 
They then compare the ranking of these taggers on the pre-defined split with their rankings %that
%obtained from 
%when 
%evaluating
%the same taggers %were evaluated 
on multiple randomly generated  splits of the same dataset. %Their results demonstrated 
The study reveals
noticeable inconsistencies in model rankings between the standard split and random splits. As a result,
the authors 
%, leading to the recommendation 
recommend adopting random splits for when comparing the performance of different model architectures.
%that the latter should be adopted for model comparisons.

\citet{sogaard-etal-2021-need} counterargue the proposal by \citet{gorman-bedrick-2019-need}.
With six tasks in English ranging from POS tagging to news classification, \citeauthor{sogaard-etal-2021-need}\ illustrate that random splits 
%tend to 
over-estimate individual model performance when it comes to new in-domain data (new test samples). By contrast, more reliable numerical estimates are obtained by adversarial splits, which partition a dataset to ensure
%such that 
%the distribution of the test set 
the test set distribution is as different as possible from that of the training set.
%, instead yield more reliable numerical estimates.

In a study that compares %comparing different 
various
data partitioning strategies for
%evaluations of 
automatic speech recognition evaluation, 
%models, 
\citet{liu-etal-2023-investigating} show that random splits, rather than adversarial splits, 
%are able to present 
offer
a more comprehensive 
capability assessment
%picture for measuring 
%of the capabilities of a 
for a
given acoustic model architecture.  This finding is particularly relevant when considering
%across 
five indigenous endangered languages with minimal training resources.

Collectively, it is not clear, based on existing findings,  which data split strategies are more capable of yielding models with more generalizable performance.
We consider that the lack of consensus among prior studies is largely due to the lack of thorough  experimentation pertaining to the number (and types) of languages, datasets and splits, as well as model architectures employed.
This study tackles these limitations by providing a sequence of data-driven experiments, with the goal of providing \textit{empirical} evidence for the capabilities of different  data partitioning strategies. 

\noindent \textbf{Morphological segmentation} Morphological segmentation has received considerable interest in the literature. Previous studies have demonstrated that incorporating morphological information 
%is effective at eliminating 
effectively eliminates data sparsity issues for a variety of downstream NLP tasks. These tasks include but are not limited to automatic speech recognition for languages such as Vietnamese ~\citep{le2009automatic} and Finnish and Turkish~\citep{kurimo2006unsupervised}, as well as machine translation for various language pairs (e.g., English $\rightarrow$ Finnish~\cite{clifton-sarkar-2011-combining}; Raramuri/Shipibo-Konibo $\leftrightarrow$ Spanish~\cite{mager-etal-2022-bpe}).

\section{Experiments}
\label{sec:experiments}

This section introduces our experimentation to investigate the impact of different data partitioning strategies on model generalizability for morphological segmentation.
We first present the details regarding the data used in our experiments, including the languages/families contained therein. We then explore the dataset construction process and describe the model architectures applied.

\subsection{Data sources}
\label{datasources}

We adopt morphological segmentation data for a total of 19 languages, spanning 10 language families, to join our experiments. 
Table~\ref{descriptive} provides relevant descriptive information and the prior works that synthesize the morphological segmentation data for the languages. Below we introduce the original data sources for each language~\cite{bender-friedman-2018-data}.
Among these languages, eight are polysynthetic indigenous/endangered Mexican languages (Mexicanero, Nahuatl, Yorem Nokki, Wixarika, Raramuri, Popoluca, Tepehua), which are all from the Yuto-Aztecan language family, and Shipibo-Konibo, which is from the Panoan language family primarily spoken in Peru and Brazil. The Raramuri data originally come from work by~\citet{caballero2010scope} 
%(including data from the associated dissertation). \bonnieshort{Seemed awkward to give same citation twice; please check my change.}
and a dissertation~\cite{caballero2008choguita}.
%The data for 
The Shipibo-Konibo data are (largely) taken from a dependency treebank~\cite{vasquez-etal-2018-toward}. 
%For the other Mexican languages, their m
Morphological segmentation data for Mexican languages are digitized from %collections of 
the Archive of Indigenous Language. % \cite{}. \bonnieshort{Does a citation go here?}
 
Seneca and Hupa are critically endangered Native American languages from the Iroquoian and Dene/Athabaskan language family respectively; the former is primarily spoken in New York State and Ontario, while the latter is the ancestral language of the Hoopa Valley
Tribe in Northern California.
%from New York and California respectively
%Seneca and Hupa are considered critically endangered languages of North America, with the former from the Iroquoian language family, while the latter is from the Dene/Athabaskan language family. 
%The data for Seneca is 
Seneca data are digitized from a grammar book~\cite{bardeau2007}, while %the data for Hupa consists 
Hupa data consist
of examples from several archival collections~\cite{curtin,kroeber,woodward1953}, along with words taken from ongoing fieldwork with an elder from the Hupa speech community. Both languages have polysynthetic morphological properties.

Next, we have two fusional, Indo-European languages: (1) English data come 
%from the Indo-European language family, English and German. The data for English is 
from the Morpho Challenge shared task for unsupervised approaches to morphological segmentation~\cite{kurimo-etal-2010-morpho}; (2)
%whereas 
%the data for German is 
German data are
harnessed from the CELEX lexical database~\cite{baayen1996celex}.

The remaining seven languages are %all 
agglutinative. 
%The data for Finnish  and Turkish  
Finish and Turkish data
%also comes 
come from the Morpho Challenge. 
Indonesian data 
%For Indonesian, the words are taken 
are from an Indonesian-English bilingual corpus.\footnote{\url{https://github.com/desmond86/Indonesian-English-Bilingual-Corpus}} 
%for Zulu, the data is 
Zulu data are collected from the Ukwabelana Corpus~\cite{spiegler-etal-2010-ukwabelana}. Lastly, the morphological segmentation data for Akan, Swahili, and Tegulu 
%is a result 
come from efforts in
%of 
the DARPA Low Resource Languages for Emerging
Incidents (LORELEI) Program~\cite{mott-etal-2020-morphological}.

%Table~\ref{descriptive} summarizes prior studies that introduce the initial data for each language, providing relevant descriptive information. We refer readers to the original studies for additional details concerning data sources and collection processes. 

% \textcolor{red}{Use ditto marks rather than blank cells so that it doesn't look like data is missing https://en.wikipedia.org/wiki/Ditto\_mark}

\subsection{Data partitioning strategy}

We explore two different data partitioning strategies: \textbf{random} and \textbf{adversarial}. Random splits divide a dataset into training and test data randomly, whereas adversarial splits partition the dataset such that the Wasserstein distance~\cite{arjovsky2017wasserstein,sogaard-etal-2021-need} between the morpheme distributions of the resulting training and test data is maximized. The aim of employing adversarial splits is to create test data that are as distant or \textit{different} as possible from the training data. Thus, model training on adversarial splits may pose greater challenges compared to random splits.

\begin{table*}
\footnotesize
\centering
\begin{tabular}{lccccc}
\hline
\textbf{Language} & \textbf{Language family} & \textbf{Morphological} & \textbf{\# word type} & \textbf{Ave. morph}  & \textbf{Data available by} \\
& & \textbf{system} & & \textbf{len} & \\
\hline
Mexicanero & Yuto-Aztecan & Polysynthetic & 882 & 3.93 &  \citealt{kann-etal-2018-fortification} \\
Nahuatl & \textquotedbl & \textquotedbl & 1,096 & 3.90 & \textquotedbl \\
Yorem Nokki & \textquotedbl & \textquotedbl & 1,050 & 3.61 & \textquotedbl\\
Wixarika & \textquotedbl & \textquotedbl &1,350 & 3.26 & \textquotedbl \\
Raramuri & \textquotedbl & \textquotedbl & 914 & 3.57 & \citet{mager-etal-2022-bpe} \\
Popoluca & \textquotedbl & \textquotedbl & 898 & 4.31 &  \citet{mager-etal-2020-tackling} \\
Tepehua & \textquotedbl & \textquotedbl & 816 & 5.37 & \textquotedbl  \\ 
Shipibo-Konibo & Panoan & \textquotedbl  & 1,096 & 4.15 & \citet{mager-etal-2022-bpe} \\
Seneca & Iroquoian & \textquotedbl & 5,425 & 2.98 & \citet{liu-etal-2021-morphological} \\ 
Hupa & Dene/Athabaskan & \textquotedbl & 595 & 3.99 & \citet{curtin} \\
&  & & & & \citet{kroeber}; \\
&  & & & & \citet{woodward1953}; \\
& & & & & linguistic fieldwork  \\
%& & & & & fieldwork with the Hupa  \\
%& & & & & speech community \\
English &  Indo-European & Fusional & 1,686 & 4.09 & \citet{cotterell-etal-2015-labeled} \\
German & \textquotedbl & \textquotedbl & 1,751 & 3.82 & \textquotedbl  \\
Finnish & Uralic & Agglutinative  & 1,835 & 4.03 & \textquotedbl  \\
Turkish & Turkic & \textquotedbl & 1,763  & 3.38 & \textquotedbl\\
Indonesian & Austronesian & \textquotedbl & 3,500 & 4.98 & \textquotedbl \\
Zulu & Niger-Congo & \textquotedbl & 10,040 & 2.37 & \textquotedbl \\
Akan & \textquotedbl & \textquotedbl & 2,046 & 2.49 & \citet{mott-etal-2020-morphological} \\
Swahili & \textquotedbl & \textquotedbl & 2,023 & 2.91 & \textquotedbl \\
Telugu & Dravidian & \textquotedbl & 2,007 & 4.07 & \textquotedbl \\
%Tegulu & & & & \\
\hline
\end{tabular}
\caption{\label{descriptive}
Descriptive statistics for the initial  morphological segmentation data of each language in our experiments; \textit{Data available by} refers to the prior work that makes the initial morphological segmentation data of the corresponding language(s) available (with the exceptions of Hupa).}
\end{table*}

\subsection{Dataset construction}
\label{construction}

\begin{figure}
    \centering    \includegraphics[width=0.3\textwidth,height=1.75in]{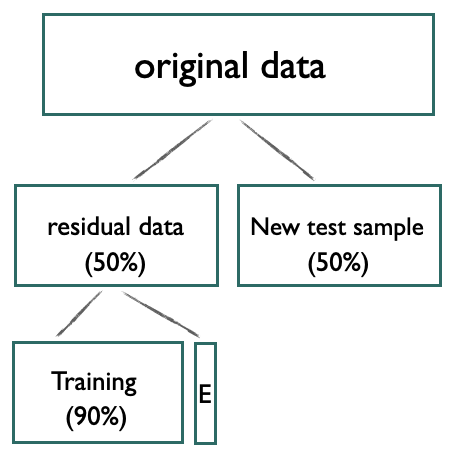}
    \caption{Simple illustration of a single dataset construction process for our experiments: given the original dataset of a language, a new test sample is constructed either randomly or adversarially (in this case, the new test sample accounts for 50\% of the original dataset); the residual data from the original data is then divided into a training and an evaluation set (\textit{E}) at a fixed 9:1 ratio, via random or adversarial splits.}
    \label{fig:datasplit}
\vspace*{-.2in}
\end{figure}

Dataset construction proceeds as follows 
%The process of constructing datasets for each language is as follows 
(see also Table~\ref{descriptive} and Figure~\ref{fig:datasplit}).
From the initial data of a given language, we first select all the unique words. The main motivation for this choice is that in practice, if a word in the test data is already included in the training data, then its morphological segmentation annotations can be directly copied from its annotations in the training data.
We refer to the resulting dataset that includes all the unique words as the \textit{original dataset} in our experiments.

%Unlike random split and adversarial split, heuristic split can only be applied once with the residual data (for each test set), due to its specific nature. However, as it turns out, there seems to be no threshold across the original datasets and test set sizes for the languages investigated.
%Therefore in what follows, we focus on results derived from random and adversarial splits. 

We determine a range of \textit{new test sample} sizes: \{10\%, 20\%, 30\%, 40\%, 50\%\} of the original dataset. Given each size, we \textit{randomly} partition the original dataset into a new test sample and a \textit{residual dataset},
%\footnote{One might wonder about whether cross-validation is applicable here. We consider random splits as similar to cross-validation,  with the following distinction: the training sets from the former possibly have overlap, whereas those from the latter do not.  We choose random splits for our experimentation based on prior work (see Table~\ref{priorwork}). However, we acknowledge that there is value in comparing random splits to cross-validation in future work.}
 10 times. In other words, for each size, we construct 10 new test samples. We consider the new test samples as approximations of new unseen data that is aside from what would normally be included in a typical dataset for practices of model training and evaluation. Ideally, one would use perhaps data of different domains to be new unseen data. That said, due to data availability limitation, we are not able to find datasets covering separate domains for the languages studied here. 
%\bonnieshort{I wonder if Eugene Li's question could be relevant here, i.e., comparison to X-validation--perhaps a footnote if room?}

For each new test sample, we use the corresponding residual dataset to build training and evaluation (eval) sets via each of the two data partitioning strategies of interest: random and adversarial splits.\footnote{We also explored \textbf{heuristic} splits as a third strategy in initial experiments. These splits are determined by considering the average morpheme count and length. In our automated search for a metric threshold in the residual data, to divide it into training and eval sets, we identify words with an average number of morphemes equal to or greater than this threshold and assign them to the eval set. 
However, we find that for most of our experimental setups, no such threshold exists.} Given each data partitioning strategy, we split the residual dataset three times, aiming for a 9:1 ratio between the resulting training and eval sets each time. As such, each individual new test sample is paired with 3 training-eval sets from random splits and another three from adversarial splits. This enables a direct comparison of models trained using the two data partitioning strategies, allowing us to determine which strategy leads to models with better performance on new test samples.

Moreover, this approach results in a total of 30 combinations consisting of a training set, an eval set, and a new test sample for each combination of a new test sample size and data partitioning strategy. 
Doing so motivates us to explore the variability across different datasets of the same size and ensure the generalization of our observations.

Lastly, we repeat the full process described above, except that this time the new test samples are generated \textit{adversarially} (i.e., partitioning the original dataset into new test samples and their corresponding residual dataset via adversarial splits). 

We note that our focus on data partitioning strategies pertains to \textit{how the residual dataset is divided}; the reason we employ different ways of generating new test samples (randomly and adversarially) is solely to see if observations will hold qualitatively, regardless of how the new test samples are derived.

\subsection{Model architectures}
\label{architectures}

We employ %\hl{four model alternatives} 
four model alternatives
from two broad model classes: conditional random field (CRF)~\cite{lafferty2001conditional},  and neural sequence-to-sequence (seq2seq) models. CRF models are 
%a type of 
log-linear discriminative models that treat
%, poses 
morphological segmentation as a sequence tagging task.  We experiment with  first-order CRF.
Given a word $w$, we add a start (<w>) and an end (</w>) symbol. 
%for which the training process involvesthe following steps:\newline
%(1) Given a word $w$, %we first appended to it 
%add a start (<w>) and an end (</w>) symbol.
%\newline
%(2) 
%Then 
For each character $w_{i}$ in the word, where $i$ represents the character's 
%marking the 
index position,
%of the character, we assigned 
we assign 
%the character with 
one of six labels: \texttt{START} (for <w>), \texttt{END} (for </w>), \texttt{S} (for any single-character morpheme), and \texttt{B} (beginning), \texttt{M} (middle), or \texttt{E} (end) for characters within a multi-character morpheme. 
As an illustration, the word \textit{avocados} will have the following sequence of segmentation labels:  
\begin{center}
\begin{small}
\begin{BVerbatim}
<w>    a v o c a d o s  </w>       
START  B M M M M M E S  END
\end{BVerbatim}
\end{small}
\end{center}
%(3) 
Lastly, for each character $w_{i}$ in $w$, we curate a feature set 
from
%composed of 
local $n$-gram (sub-)strings,
%. This feature set is used %which 
as input to a first-order CRF model, 
%used 
to predict the corresponding label %of 
for $w_{i}$. All CRF models 
%were 
are implemented 
using the
%in 
\texttt{python-crfsuite} framework.\footnote{\url{https://python-crfsuite.readthedocs.io/en/latest/}}

For seq2seq, we use \texttt{fairseq}~\cite{ott-etal-2019-fairseq} to explore three different encoder-decoders:
%, including 
\textsc{LSTM}, \textsc{Transformer}, and \textsc{Transformer}\_\textsc{tiny}.\footnote{\url{https://fairseq.readthedocs.io/en/latest/models.html}} For each encoder-decoder architecture, the model input is always the word itself as a sequence of letters (with space between every two consecutive letters), and the model output contains an extra exclamation point (!) as indication of morpheme boundary.
%All seq2seq models are implemented %in \texttt{fairseq} 
%using the default parameters, trained with 3 unique random seeds.

\begin{center}
\begin{small}
\begin{BVerbatim}
INPUT    a v o c a d o s  
OUTPUT  a v o c a d o ! s
\end{BVerbatim}
\end{small}
\end{center}

\noindent All seq2seq models are implemented using the default parameters: 
for the \textsc{LSTM}-based architecture, all embeddings have 512 dimensions; both the
encoder and the decoder contain one hidden layers
with 512 hidden units in each layer; \textsc{Transformer} has 6 encoder-decoder layers, 8 self-attention heads, an embedding size of 512, and 2048 hidden units in the feed-forward layers; \textsc{Transformer\_tiny} has 2 encoder-decoder layers, 2 self-attention heads, with the embedding dimension and feed-forward layer dimension both being 64.

In all experimental configurations conducted in this study,
%Across all the experimental settings here, 
the parameter implementation for each model architecture 
%is 
remains
the same for all languages (see also Appendix~\ref{sec:computing}). The model evaluation is performed using %We use 
the $F1$ score as the 
%model evaluation
metric.

\section{Analysis}
\label{sec:analysis}

Our analysis seeks to address %the following 
two
%following 
questions:

\hspace*{-.25in}
\begin{tabular}{lp{2.65in}}
    (1)&Which data partitioning strategy leads to more accurate numerical ``guesses" of, as well as better, individual model performance for the new test samples?\\
    (2)&How do different data partitioning strategies affect the generalization of model rankings from the eval sets to new test samples?
%    \item How much variation is there in model performance across new test samples with the same size given each data partitioning strategy?
%    \item What contributes to the variation in model performance for both data partitioning strategies?
\end{tabular}

%\textcolor{blue}{The charts are nice, but please specify the axes, i.e., state what are 0, .05, .10, etc. and/or include in the caption (Maybe use "Avg F1 score difference" and in caption refer to it in the long form: "Average F1 score difference between eval sets and new test samples"). Also, convey somewhere that a high number is bad. :-) Below I added "where a \textit{small} F1 score is considered better." -- make sure this wording is okay.}

Note that for each research question, we perform analysis of each individual language first, then focus on a summary of aggregated averages across languages; languages with idiosyncratic patterns, however, are noted when necessary.
Throughout our analysis, we first describe results from cases where the new test samples are derived randomly. We then move onto settings where the new test samples are generated adversarially, in order to see if there are notable similarities and differences in the observations between the two.

\subsection{Individual model performance}
\label{individual}

This section analyzes estimates of individual model performance using different data partitioning strategies (applied to residual data), \textit{when the new test samples are generated randomly}. Given each language, for every model architecture, we measure the average $F1$ score difference between the eval sets and their corresponding new test samples. A higher average $F1$ score difference indicates that the performance of a given model does not generalize well from the eval sets to new test samples.

Across the 19 languages, 
%\bonnieshort{I struggled to understand what you meant by former and latter in this first sentence--I edited a bit, but am worried I have no introduced an error; please check and fix.}
adversarial splits lead to much larger score differences for all four models, with the scores for new test samples
%for the former 
consistently higher than those for eval sets (Table~\ref{individualmodel}).
%the latter.
On the contrary, for random splits, there is no noticeable score difference for any of the model architectures. (Detailed language-by-language results are in Table~\ref{detail_random} in Appendix~\ref{sec:appendix_detailed}).
Collectively, these patterns suggest when focusing solely on the achieved scores of a model, random splits provide more reliable numerical estimates compared to adversarial splits.

\begin{table}[h!]
\footnotesize
\resizebox{\columnwidth}{!}{
\centering
\begin{tabular}{llll}
\hline
\textbf{Model} & \textbf{Split} & \textbf{Eval sets} & \textbf{New test} \\\hline  
\textsc{CRF} & random & 0.80 & 0.80 \\
& adversarial & 0.76 & 0.79 \\
\textsc{TRM\_tiny} & random & 0.68 & 0.68 \\
& adversarial & 0.59 & 0.67 \\
\textsc{LSTM} & random & 0.67 & 0.67 \\
& adversarial & 0.62 & 0.65 \\
\textsc{TRM} & random & 0.56 & 0.56 \\
& adversarial & 0.48 & 0.54 \\
\hline
\end{tabular}
}
\caption{
Individual model performance ($F1$) for eval sets and new test samples averaged across languages, when the new test samples are generated \textbf{randomly}.
%; \textsc{TRM} stands for \textsc{Transformer} and \textsc{TRM\_Tiny} for \textsc{Transformer\_Tiny}
}
\label{individualmodel}
\end{table}

Furthermore, we compare which data partitioning strategy results in higher scores for the new test samples. As shown in Table~\ref{individualmodel}, random splits consistently lead to (slightly) better model performance across the four model architectures. Among the different models, the 
largest performance gap from the two data partitioning strategies is 
observed for \textsc{LSTM} (0.02) and \textsc{transformer} (0.02). Similar observations exist when analyzing variously sized training sets and new test samples.

We carry out the same analysis for cases \textit{where the new test samples are derived adversarially} (Table~\ref{individualmodel_adversarial}). We find that adversarial splits of the residual data \textit{instead} lead to lower $F1$ score difference (0.09) on average across settings, compared to random splits (0.15). This holds mostly when breaking down by languages and individual model architectures as well. That said, randomly partitioning the residual data yields better average model performance for new test samples. (See Table~\ref{detail_adversarial} in Appendix~\ref{sec:appendix_detailed} for language-by-language results.) These results suggest that if one were to care mainly about achieving a higher numerical score on additional data with a given model architecture, random splits would be a more suitable option.

\begin{table}[h!]
\footnotesize
\resizebox{\columnwidth}{!}{
\centering
\begin{tabular}{llll}
\hline
\textbf{Model} & \textbf{Split} & \textbf{Eval sets} & \textbf{New test} \\\hline  
\textsc{CRF} & random & 0.80 & 0.65 \\
& adversarial & 0.67 & 0.59 \\
\textsc{TRM\_tiny} & random & 0.69 & 0.52 \\
& adversarial & 0.55 & 0.46 \\
\textsc{LSTM} & random & 0.66 & 0.52 \\
& adversarial & 0.55 & 0.45 \\
\textsc{TRM} & random & 0.55 & 0.41 \\
& adversarial & 0.44 & 0.37 \\
\hline
\end{tabular}
}
\caption{
Individual model performance ($F1$) for eval sets and new test samples averaged across languages, when the new test samples are generated \textbf{adversarially}.
%; \textsc{TRM} stands for \textsc{Transformer} and \textsc{TRM\_Tiny} for \textsc{Transformer\_Tiny}
}
\label{individualmodel_adversarial}
\end{table}

\subsection{Model ranking}
\label{ranking}

We now turn to studying the effect of different data partitioning strategies on model ranking generalizations, \textit{when the new test samples are derived randomly}. For each of the two data partitioning strategies, given every combination of a training set, an eval set, and a new test sample, we derive the ranking of the four model architectures based on their F1 scores (averaged across 3 random seeds) on the eval set ($Ranking\_eval$) and the new test sample ($Ranking\_new$), respectively. (Again, based on how we construct the datasets initially, the two $F1$ scores are predicted by the same model from the training set, thereby directly comparable). 
%We present findings for cases where the new test samples are generated randomly. (Similar observations persist when the new test samples are derived adversarially.)

\textbf{Best overall model ranking} We compute the best overall model ranking (e.g., the most frequent
%which 
$Ranking\_eval$)
%is the most frequent) 
for both the eval sets and the new test samples, considering each data partitioning strategy. 
Comparing results across languages, 
%it appears that 
for both random and adversarial partitions, the best overall ranking is \textsc{CRF} $>$ \textsc{Transformer\_Tiny} $>$ \textsc{LSTM} $>$ \textsc{Transformer}. This ranking holds for twelve out of the 19 languages examined here (Table~\ref{ranking_results}), including examples from all three morphological systems covered in this study.
Additionally, for these languages, there are on average  noticeable $F1$ score differences between \textsc{CRF} (the best model) and \textsc{Transformer\_Tiny} (the second best model) for both eval sets (random: 0.12; adversarial: 0.16) and new test samples (random: 0.12; adversarial: 0.12).

\begin{table}[h!]
%\footnotesize
\resizebox{\columnwidth}{!}{
\centering
\begin{tabular}{l}
%\hline
\textbf{\textsc{CRF} $>$ \textsc{Transformer\_Tiny} $>$ \textsc{LSTM} $>$ \textsc{Transformer}}   \\\hline
Mexicanero, Nahuatl, Yorem Nokki, Raramuri, Popoluca,  \\
Shipibo-Konibo, Hupa, English, Turkish, Indonesian, Swahili, Telugu \\
% \hline
 \\
\textbf{\textsc{CRF} $>$ \textsc{LSTM}$>$ \textsc{Transformer\_Tiny}  $>$ \textsc{Transformer}} \\\hline
Wixarika, Tepehua, Seneca, German, Finnish,  Zulu \\
%\hline 
\\
\textbf{\textsc{CRF} $>$ \textsc{LSTM}  $>$ \textsc{Transformer} $>$ \textsc{Transformer\_Tiny}} \\\hline
Akan \\
%\hline
\end{tabular}
}
\caption{
Results for the best overall rankings, when the new test samples are generated \textbf{randomly}.}
\label{ranking_results}
\end{table}

The best overall ranking above is followed by an alternative 
%a ranking 
that is the best for six other languages (\textsc{CRF} $>$ \textsc{LSTM}$>$ \textsc{Transformer\_Tiny}  $>$ \textsc{Transformer}), covering different morphological properties as well (Table~\ref{ranking_results}). 
%\bonnieshort{Do you want to comment on properties of the 6 languages that set them apart, if room?  (Again, maybe a footnote?)}
The main difference between the two rankings pertains to \textsc{LSTM} and \textsc{Transformer\_Tiny}. 
%In terms of 
For languages where the second best overall ranking applies, the average $F1$ difference between \textsc{LSTM} and \textsc{Transformer\_Tiny} is mostly smaller than 0.02. 
Again, there are notable average $F1$ differences 
%in average $F1$ scores 
between \textsc{CRF} and the second best performing model, \textsc{LSTM}, for eval sets (random: 0.12; adversarial: 0.18) and new test samples (random: 0.12; adversarial: 0.12).

%First, for both random and adversarial partitions, the best overall ranking across languages appears to be: \textsc{CRF} $>$ \textsc{Transformer\_Tiny} $>$ \textsc{LSTM} $>$ \textsc{Transformer} (Table~\ref{ranking}).
%The observed trend indicates that for both random and adversarial splits, the best overall ranking remains consistent for most of the languages between the eval sets and the new test samples: \textsc{CRF} $>$ \textsc{Transformer\_Tiny} $>$ \textsc{LSTM} $>$ \textsc{Transformer}.

%There is on average a noticeable $F1$ score difference between \textsc{CRF} and \textsc{Transformer\_tiny}, with the latter demonstrating the second-best overall performance (see also Table~\ref{individualmodel}). 
%The only three exceptions here are Turkish, Indonesian, and Akan, for which the best model ranking is \textsc{CRF} $>$ \textsc{LSTM}$>$ \textsc{Transformer\_Tiny}  $>$ \textsc{Transformer}.
%For Indonesian, there is a minor score difference between \textsc{CRF} (0.86) and \textsc{Transformer\_Tiny} (0.87), whereas the score discrepancy is larger for Turkish (\textsc{CRF}: 0.64; \textsc{Transformer\_Tiny}: 0.68).

\textbf{Overall model ranking generalizability} %In addition, we 
We also examine model ranking consistency between each eval-set/new-test pair by
%. To that end, we measure 
measuring
%ing 
the proportion of cases where $Ranking\_new$ is the same as $Ranking\_eval$ for training/eval/new-test combinations. %a training set, an eval set, and a new test sample. 
Given that the $F1$ score difference between \textsc{LSTM} and \textsc{Transformer\_Tiny} is relatively minimal (see above and Table~\ref{individualmodel}), we collapse the two best rankings described above 
%(\textsc{CRF} $>$ \textsc{Transformer\_Tiny} $>$ \textsc{LSTM} $>$ \textsc{Transformer} and \textsc{CRF} $>$ \textsc{LSTM}$>$ \textsc{Transformer\_Tiny}  $>$ \textsc{Transformer}) 
when measuring model ranking consistency.
On average,
the 
%proportion of cases where the model ranking is 
consistency of model rankings
%the same for both the 
between eval sets and the new test samples ($Ranking\_eval$ = $Ranking\_new$) is  higher for random  splits (91.47\%) than  for adversarial splits (90.26\%).
This pattern  persists for most of the languages.% (Table~\ref{rankingconsistency}).

%\textbf{Model ranking variability} A question then arises: how much variability is there in model rankings for the two partitioning strategies for new unseen data? We demonstrate our quantification of ranking variability below, using Yorem Nokki as an example. Across all the eval sets of random splits, for instance, there are five different model rankings; we measure the proportion of each model ranking ($P\_ranking$), then use entropy~\citep{cover1991information} to compute the amount of variability in model rankings (Eq. (1)).
%\begin{equation}
%    -\sum P\_rankings * log P\_rankings
%\end{equation}
%Our results show that on average across all languages, for the eval sets, there is minimal difference in entropy value between random (1.10) and adversarial splits (1.07), suggesting that there is comparable amount of variation in model rankings between the two data partitioning strategy. These observations hold, qualitatively, for the new test samples as well (random: 0.88; adversarial: 0.84).

We now investigate settings with \textit{adversarially generated new test samples}. The two best overall model rankings %appear to be 
are the same as the cases above, where
%when the 
new test samples are constructed randomly. 
%With regards to 
Regarding the consistency of model rankings, again, we merge the two best overall rankings. %together. 
The results show that the average proportion of cases where $Ranking\_new$ and $Ranking\_eval$ are the same is higher for random splits (78.67\%) in contrast to observations from adversarial splits (75.79\%). The average numerical discrepancy here (2.88\%) is also larger than what is reported above for randomly constructed new test samples (91.47\%-90.26\%=1.21\%).
These findings indicate random splits possibly
%are able to 
yield more reliable model ranking results in the face of
%when facing 
new test samples.

%Our results show that consistently, there is more variability in model ranking derived from random splits for the new test samples (0.27), in comparison to observations from adversarial splits (0.16). These observations hold, qualitatively, across differently sized combinations of training, eval sets and new test samples. 
%On average across all languages, we find observations similar to those observed for Yorem Nokki, where the amount of variation in model ranking is also higher for random (0.52) than adversarial splits (0.44).
%Taken together, these findings suggest that adversarial splits instead tend to yield better model ranking generalization,

\subsection{Variation across datasets}

%\zoey{Would this section make sense? There did not turn out to be very high score variability (cf. Figure 2), which otherwise will make a more interesting story; but I think it would be worthwhile to resport this?}

Thus far, our analysis focuses on scores averaged across datasets. Recall that, for each language and each combination of 
%, with each pair of a 
new test sample size and data partitioning strategy, we construct 30 
%combinations 
sets comprised of a training set, an eval set, and a new test sample.
This section aims to better understand the extent of variability in model performance  
between the two data partitioning strategies across new test samples of the same sizes. This can, in turn, shed light on the reliability of our prior analysis, which depends on average scores across
%that relies on averaged scores in 
different settings.

\begin{figure}[h!]
\vspace*{-.15in}
    \centering    \includegraphics[width=0.44\textwidth,height=2.5in]{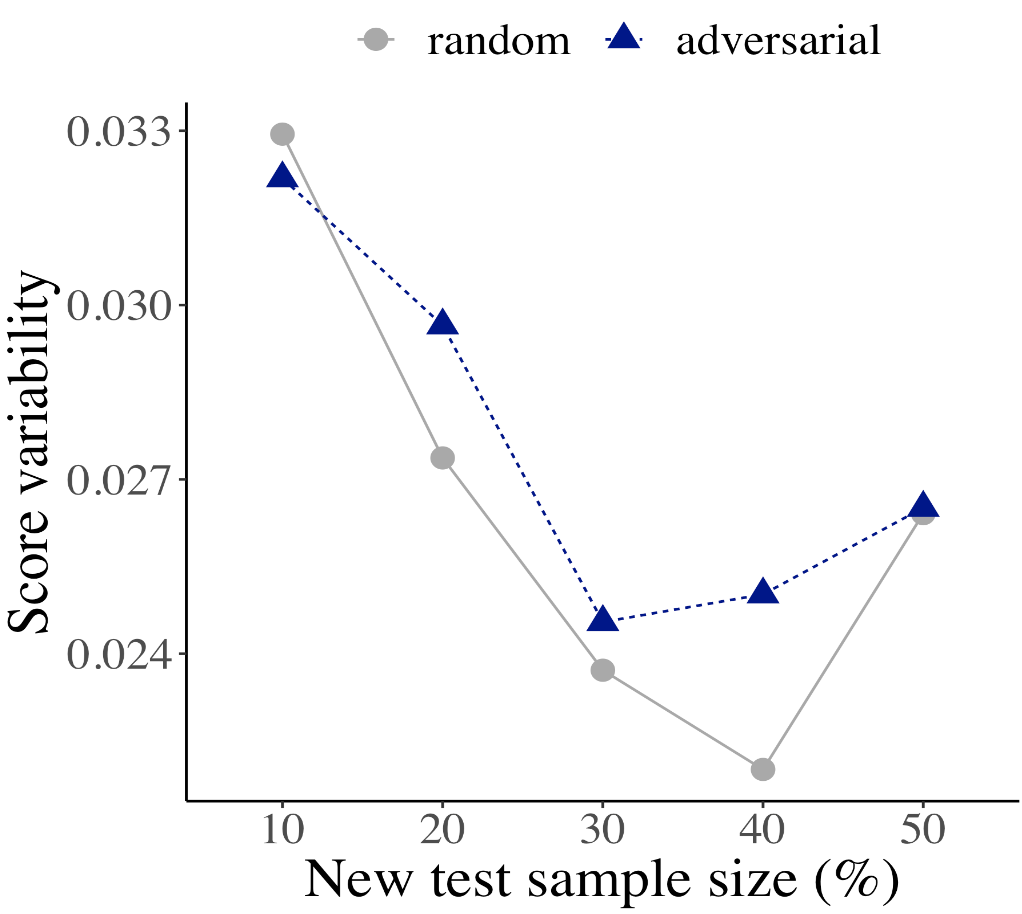}
    \caption{Score variability for every eval set size (\%) given each data partitioning strategy averaged across languages and model architectures, when the new test samples are generated \textbf{randomly}.}
    \label{fig:size_score_variability}
\end{figure}

%To address this, with 
For each new test sample size
%pair of a new test sample size 
and data partitioning strategy (for a given language), we calculate %the 
$F1$ 
%score 
variability (standard deviation) for each model architecture across the 30 combinations of a training set, an eval set, and a new test sample. We then measure the average score variability across languages and model architectures. As shown in Figure~\ref{fig:size_score_variability}, when the new test samples are constructed via random sampling, the average score variability exhibits comparable values  for almost all new test sample sizes between random and adversarial splits.  The score variability values predominantly fall below 0.03; this suggests that there is a small amount of variation across datasets with the same new test sample sizes.  Compared to the other three neural alternatives, \textsc{CRF} exhibits the least
%amount of 
performance variation; across model architectures,  \textsc{Transformer\_tiny} demonstrates the highest $F1$ variability for both data partitioning strategies.

%we calculate two measures for each combination of language, model, new test sample size, and data partitioning strategy: 
%(1) \textit{score range}, which represents the difference between the highest and the lowest $F1$ score; and (2) \textit{score variability}, which indicates the standard deviation of all the $F1$ scores.
%obtained for new test samples of a given size.

%Let's again consider the example of Yorem Nokki. 

%As presented in 

Similar observations are found for cases where the new test samples are constructed adversarially.
These patterns further validate our previous findings on individual model performance and model rankings averaged across datasets.

\subsection{Regression analysis}

Having established that random splits lead to comparatively better individual model performance and more generalizable model rankings on new test samples, this section aims to add statistical rigor to our findings. To achieve this, we resort to regression analysis. Our variable of interest is the data partitioning strategy applied to the residual data, which has 
interaction terms with %\bonnieshort{I did some "fat trimming" in this paragraph, and probably botched it. I also had trouble understanding what "interaction terms". So please check this paragraph carefully for errors.}
%each of the following 
various control variables:
the method of deriving
%: the way the new test samples are derived 
new test samples
(randomly or adversarially), morpheme overlap (the proportion of morphemes in the eval set that 
%also 
occur in the training set), and the relative ratios between training and eval sets for the average number of \textit{morphemes} per word and the average number of morpheme \textit{types} per word.
%the number of words, the average number of morphemes per word, and the average number of morpheme types per word. 
Lastly, we control for the model architecture applied.

Ideally, we would fit one mixed-effect regression model including all factors described above, with the language as a random effect. %Given that the 
The full dataset 
%containing results 
resulting from all experiments, however, is quite large ($N$ = 91,200); therefore we turn to fitting one linear regression model for each language instead. The goal here is to 
determine whether the superior performance of random split
%see whether  that random split 
\textit{is an observation that can be found for all, or most of the languages, regardless of their respective dataset size}.

\begin{table}[h!]
\footnotesize
%\resizebox{\columnwidth}{!}{
\centering
\begin{tabular}{ll}
\hline
\textbf{Language} & \textbf{Data partitiong strategy $\beta$} \\\hline
Mexicanero & 0.05*** \\
Nahuatl &  0.34*** \\
Yorem Nokki & 0.67*** \\
Wixarika & -0.30*** \\
Raramuri & 0.42*** \\
Popoluca & 0.31*** \\
Tepehua & 0.08** \\
Shipibo-Konibo & 0.80*** \\
Seneca & 1.12*** \\
Hupa & 0.14*** \\
English & 0.33*** \\
German & -0.19*** \\
Finnish & 0.50*** \\
Turkish & 0.46*** \\
Indonesian & 0.29*** \\
Zulu & 1.90*** \\
Akan & 0.26 \\
Swahili & 0.20* \\
Telugu & 0.10** \\
\hline
\end{tabular}
%}
\caption{
Regression coefficients ($\beta$) of data partitioning strategy for each language; a positive coefficient value indicates that randomly splitting the residual data has a positive effect on $F1$ scores, while a negative value denotes the opposite; the number of * suggests significance level: * $p < 0.05$, ** $p < 0.01$, *** $p < 0.001$.}
\label{regression}
\end{table}

The regression coefficients for the data partitioning strategy are presented in Table~\ref{regression} (see Table~\ref{regression_full} in Appendix~\ref{sec:regression_full} for coefficients of other control variables). A significantly positive coefficient value means  random splits lead to significantly higher $F1$ scores, in contrast to adversarial splits. This pattern is evident in 16 out of nineteen languages, with the exceptions of Wixarika and German, where random splits appear to have a significantly negative effect on model performance, and Akan, where there is no pronounced difference between the two data partitioning strategies. The statistics overall further corroborate our prior observations.

\section{Conclusions}
\label{sec:discussion}

This study investigates the impact of data partitioning strategies on model generalizability, using morphological segmentation as a test case, drawing data from 19 typologically diverse languages, including ten indigenous/endangered languages. 
Our results demonstrate that, independent of the morphological properties of the languages, random splits, in contrast to adversarial splits, yield: (1) better model performance, and (2) more reliable model rankings on %, when it comes to 
new test data. These patterns hold across varying sized combinations of training and eval sets, as well as new test samples, 
as evidenced by the minimal
%which is evident from the fact that we find very small amount of 
variation in model performance across datasets 
%of the languages understudied.
for the languages examined. The findings are also supported 
%from 
by our statistical regression analysis,
where random splits are shown to have a pronounced positive impact on model performance; this pattern holds for most languages, in spite of the fact that each individual language has a different dataset size.
% where random splits are shown to have a pronounced positive impact on model performance.

It is worth %nothing 
noting that the average $F1$ score differences between random and adversarial splits on the new test samples are much larger across model architectures, when the new test samples are generated adversarially, in comparison to when they are derived randomly (Section~\ref{individual}). What's more, the trend that random splits yield more consistent model rankings is stronger when facing adversarially constructed new test samples as well (Section~\ref{ranking}). Recall that adversarial samples are posed to be as distant from the training data as possible. These tendencies suggest that when facing more challenging new test data in the wild (challenging \textit{relative to the training data}), there is potentially more benefit in applying random splits, at least in the case of morphological segmentation.

While random splits seem to outperform adversarial splits in our study, we do not wish to draw the same conclusions for other tasks. With the methodologies outlined here, for future work, we would like to expand to different tasks from a cross-linguistic angle. In particular, we are interested in settings where the languages have \textit{a spectrum of data availability}, in order to probe what  data partitioning strategy will be preferred given different extents of data limitation. In addition, while heuristic splits are not plausible in our experiments, in future cases where applicable (e.g., text classification where the dataset can possibly be split heuristically based on the number of tokens or token types in the sentences), there is potential value in including such splits for more thorough comparisons.

\section*{Limitations}
\label{sec:limitations}

Our study faces two primary limitations.
First, as described in Section~\ref{datasources}, for each language, the data for experimentation come
%restricted availability -> limited availability (BJD: I changed this because otherwise it sounds like you're saying they are available but there are restrictions on your ability to access them.)
from the same domain, due to limited availability of datasets for morphological segmentation.

Second, since indigenous and endangered languages are often resource-constrained, after constructing each new test sample, we split the residual data into training and eval sets in order for the experimental setups to be consistent across languages, thereby not including a validation (or a tune) set for model parameter tuning. That said, the need for parameter tuning itself is an indication, to some extent, that the model may not generalize well to new data. 

%Second, since indigenous and endangered languages are often resource-constrained, to be consistent across languages, after constructing each new test sample, we only split the residual data into training and eval sets, thereby not including a validation (or a tune) set for model parameter tuning. That said, the need for parameter tuning itself is an indication, to some extent, that the model may not generalize well to new data. 

%Relatedly, how to split a given dataset, especially adversarially, into training, validation, and eval sets is not exactly clear.
%In prior work, \citet{van-der-goot-2021-need} proposes to divide a given dataset into train-tune-validation-test split for the purpose of avoiding overfitting. However, it is unclear how one would arrive at such split. The target task for \citet{van-der-goot-2021-need}'s work is dependency parsing using data from the Universal Dependencies project (UD)~\citep{de-marneffe-etal-2021-universal}. One of the benefits of the UD datasets is that all sentences are specified in a given order (e.g., a set of sentences produced by one author followed by another set produced by a different author). Therefore a  UD dataset can be split into train-tune-validation-test split sequentially. 

%By contrast, datasets for morphological segmentation are not necessarily constructed in a similar way, which makes applying a similar split less straightforward. Nevertheless, resource permitting, it does make an intriguing direction to design alternative data partitioning strategies that will take into account validation sets.

\section*{Ethics Statement and Broader Impact}

Our study  compares data partitioning strategies in order to better understand model generalizability, especially for indigenous languages. This research not only provides insights into conducting more reliable model evaluation in a broader sense, but also informs research for the development of  more effective language technology  for indigenous and endangered languages. Such advancements can contribute to language documentation efforts and support the respective speech communities. In addition, our selection of languages contributes to the ongoing efforts to promote language diversity in the field of natural language processing.

All original datasets used in our paper are publicly available, except for Hupa, which is an in-house dataset, derived with permission granted through academic relations as well as indirect relations with enthusiastic cooperation of the elders from the Hupa speech community. Therefore, ethical concerns of using the Hupa morphology data have been carefully considered.

%\bibliography{custom,anthology}

\section*{Acknowledgements}

We would like to thank Jordan Kodner for helpful feedback on earlier drafts of our paper. This material is based upon work supported, in part, by the Defense Advanced Research Projects Agency (DARPA) under Contract No. HR001121C0186.

%\bibliography{custom,anthology}

\begin{thebibliography}{37}
\expandafter\ifx\csname natexlab\endcsname\relax\def\natexlab#1{#1}\fi

\bibitem[{Arjovsky et~al.(2017)Arjovsky, Chintala, and Bottou}]{arjovsky2017wasserstein}
Martin Arjovsky, Soumith Chintala, and L{\'e}on Bottou. 2017.
\newblock Wasserstein generative adversarial networks.
\newblock In \emph{International conference on machine learning}, pages 214--223. PMLR.

\bibitem[{Baayen et~al.(1996)Baayen, Piepenbrock, and Gulikers}]{baayen1996celex}
R~Harald Baayen, Richard Piepenbrock, and Leon Gulikers. 1996.
\newblock The {CELEX} lexical database (cd-rom).

\bibitem[{Bardeau(2007)}]{bardeau2007}
Phyllis E.~Wms. Bardeau. 2007.
\newblock \emph{The Seneca Verb: Labeling the Ancient Voice}.
\newblock Seneca Nation Education Department, Cattaraugus Territory.

\bibitem[{Bender(2019)}]{bender2019benderrule}
Emily Bender. 2019.
\newblock The \#{B}ender{R}ule: On naming the languages we study and why it matters.
\newblock \emph{The Gradient}.

\bibitem[{Bender and Friedman(2018)}]{bender-friedman-2018-data}
Emily~M. Bender and Batya Friedman. 2018.
\newblock \href {https://doi.org/10.1162/tacl_a_00041} {Data statements for natural language processing: Toward mitigating system bias and enabling better science}.
\newblock \emph{Transactions of the Association for Computational Linguistics}, 6:587--604.

\bibitem[{Caballero(2008)}]{caballero2008choguita}
Gabriela Caballero. 2008.
\newblock \emph{Choguita Rar{\'a}muri (Tarahumara) phonology and morphology}.
\newblock Ph.D. thesis, University of California, Berkeley.

\bibitem[{Caballero(2010)}]{caballero2010scope}
Gabriela Caballero. 2010.
\newblock Scope, phonology and morphology in an agglutinating language: Choguita {R}ar{\'a}muri ({T}arahumara) variable suffix ordering.
\newblock \emph{Morphology}, 20:165--204.

\bibitem[{Clifton and Sarkar(2011)}]{clifton-sarkar-2011-combining}
Ann Clifton and Anoop Sarkar. 2011.
\newblock \href {https://aclanthology.org/P11-1004} {Combining morpheme-based machine translation with post-processing morpheme prediction}.
\newblock In \emph{Proceedings of the 49th Annual Meeting of the Association for Computational Linguistics: Human Language Technologies}, pages 32--42, Portland, Oregon, USA. Association for Computational Linguistics.

\bibitem[{Collins(2002)}]{collins-2002-discriminative}
Michael Collins. 2002.
\newblock \href {https://doi.org/10.3115/1118693.1118694} {Discriminative training methods for hidden {M}arkov models: Theory and experiments with perceptron algorithms}.
\newblock In \emph{Proceedings of the 2002 Conference on Empirical Methods in Natural Language Processing ({EMNLP} 2002)}, pages 1--8. Association for Computational Linguistics.

\bibitem[{Cotterell et~al.(2015)Cotterell, M{\"u}ller, Fraser, and Sch{\"u}tze}]{cotterell-etal-2015-labeled}
Ryan Cotterell, Thomas M{\"u}ller, Alexander Fraser, and Hinrich Sch{\"u}tze. 2015.
\newblock \href {https://doi.org/10.18653/v1/K15-1017} {Labeled morphological segmentation with semi-{M}arkov models}.
\newblock In \emph{Proceedings of the Nineteenth Conference on Computational Natural Language Learning}, pages 164--174, Beijing, China. Association for Computational Linguistics.

\bibitem[{Curtin(1888-1889)}]{curtin}
Jeremiah Curtin. 1888-1889.
\newblock \emph{Hupa vocabulary December 1888-January 1889}.
\newblock National Anthropological Archives: NAA MS 2063.

\bibitem[{de~Marneffe et~al.(2021)de~Marneffe, Manning, Nivre, and Zeman}]{de-marneffe-etal-2021-universal}
Marie-Catherine de~Marneffe, Christopher~D. Manning, Joakim Nivre, and Daniel Zeman. 2021.
\newblock \href {https://doi.org/10.1162/coli_a_00402} {{U}niversal {D}ependencies}.
\newblock \emph{Computational Linguistics}, 47(2):255--308.

\bibitem[{Ducel et~al.(2022)Ducel, Fort, Lejeune, and Lepage}]{DBLP:conf/lrec/DucelFLL22}
Fanny Ducel, Kar{\"{e}}n Fort, Ga{\"{e}}l Lejeune, and Yves Lepage. 2022.
\newblock \href {https://aclanthology.org/2022.lrec-1.60} {Do we name the languages we study? {T}he {\#}{B}ender{R}ule in {LREC} and {ACL} articles}.
\newblock In \emph{Proceedings of the Thirteenth Language Resources and Evaluation Conference, {LREC} 2022, Marseille, France, 20-25 June 2022}, pages 564--573. European Language Resources Association.

\bibitem[{Gauthier et~al.(2016)Gauthier, Besacier, Voisin, Melese, and Elingui}]{gauthier-etal-2016-collecting}
Elodie Gauthier, Laurent Besacier, Sylvie Voisin, Michael Melese, and Uriel~Pascal Elingui. 2016.
\newblock \href {https://aclanthology.org/L16-1611} {Collecting resources in sub-{S}aharan {A}frican languages for automatic speech recognition: a case study of {W}olof}.
\newblock In \emph{Proceedings of the Tenth International Conference on Language Resources and Evaluation ({LREC}'16)}, pages 3863--3867, Portoro{\v{z}}, Slovenia. European Language Resources Association (ELRA).

\bibitem[{Gorman and Bedrick(2019)}]{gorman-bedrick-2019-need}
Kyle Gorman and Steven Bedrick. 2019.
\newblock \href {https://doi.org/10.18653/v1/P19-1267} {We need to talk about standard splits}.
\newblock In \emph{Proceedings of the 57th Annual Meeting of the Association for Computational Linguistics}, pages 2786--2791, Florence, Italy. Association for Computational Linguistics.

\bibitem[{Kann et~al.(2018)Kann, Mager~Hois, Meza-Ruiz, and Sch{\"u}tze}]{kann-etal-2018-fortification}
Katharina Kann, Jesus~Manuel Mager~Hois, Ivan~Vladimir Meza-Ruiz, and Hinrich Sch{\"u}tze. 2018.
\newblock \href {https://doi.org/10.18653/v1/N18-1005} {Fortification of neural morphological segmentation models for polysynthetic minimal-resource languages}.
\newblock In \emph{Proceedings of the 2018 Conference of the North {A}merican Chapter of the Association for Computational Linguistics: Human Language Technologies, Volume 1 (Long Papers)}, pages 47--57, New Orleans, Louisiana. Association for Computational Linguistics.

\bibitem[{Kodner et~al.(2023)Kodner, Payne, Khalifa, and Liu}]{kodner-etal-2023-morphological}
Jordan Kodner, Sarah Payne, Salam Khalifa, and Zoey Liu. 2023.
\newblock \href {https://doi.org/10.18653/v1/2023.acl-long.335} {Morphological inflection: A reality check}.
\newblock In \emph{Proceedings of the 61st Annual Meeting of the Association for Computational Linguistics (Volume 1: Long Papers)}, pages 6082--6101, Toronto, Canada. Association for Computational Linguistics.

\bibitem[{Kroeber(1900-1906)}]{kroeber}
Alfred Kroeber. 1900-1906.
\newblock \emph{Untitled Hupa text}.
\newblock Transcription in Kroeber’s hand included in Goddard (1903-1906), notebook \#4.

\bibitem[{Kurimo et~al.(2006)Kurimo, Creutz, Varjokallio, Arsoy, and Saraclar}]{kurimo2006unsupervised}
Mikko Kurimo, Mathias Creutz, Matti Varjokallio, Ebru Arsoy, and Murat Saraclar. 2006.
\newblock Unsupervised segmentation of words into morphemes-morpho challenge 2005 application to automatic speech recognition.
\newblock In \emph{Ninth International Conference on Spoken Language Processing}, pages 1021--1024.

\bibitem[{Kurimo et~al.(2010)Kurimo, Virpioja, Turunen, and Lagus}]{kurimo-etal-2010-morpho}
Mikko Kurimo, Sami Virpioja, Ville Turunen, and Krista Lagus. 2010.
\newblock \href {https://aclanthology.org/W10-2211} {Morpho challenge 2005-2010: Evaluations and results}.
\newblock In \emph{Proceedings of the 11th Meeting of the {ACL} Special Interest Group on Computational Morphology and Phonology}, pages 87--95, Uppsala, Sweden. Association for Computational Linguistics.

\bibitem[{Kwiatkowski et~al.(2019)Kwiatkowski, Palomaki, Redfield, Collins, Parikh, Alberti, Epstein, Polosukhin, Devlin, Lee, Toutanova, Jones, Kelcey, Chang, Dai, Uszkoreit, Le, and Petrov}]{kwiatkowski-etal-2019-natural}
Tom Kwiatkowski, Jennimaria Palomaki, Olivia Redfield, Michael Collins, Ankur Parikh, Chris Alberti, Danielle Epstein, Illia Polosukhin, Jacob Devlin, Kenton Lee, Kristina Toutanova, Llion Jones, Matthew Kelcey, Ming-Wei Chang, Andrew~M. Dai, Jakob Uszkoreit, Quoc Le, and Slav Petrov. 2019.
\newblock \href {https://doi.org/10.1162/tacl_a_00276} {Natural questions: A benchmark for question answering research}.
\newblock \emph{Transactions of the Association for Computational Linguistics}, 7:452--466.

\bibitem[{Lafferty et~al.(2001)Lafferty, McCallum, and Pereira}]{lafferty2001conditional}
John Lafferty, Andrew McCallum, and Fernando~C.N. Pereira. 2001.
\newblock Conditional random fields: {P}robabilistic models for segmenting and labeling sequence data.
\newblock In \emph{Proceedings of the Eighteenth International Conference on Machine Learning}, pages 282--289.

\bibitem[{Le and Besacier(2009)}]{le2009automatic}
Viet-Bac Le and Laurent Besacier. 2009.
\newblock Automatic speech recognition for under-resourced languages: application to {V}ietnamese language.
\newblock \emph{IEEE Transactions on Audio, Speech, and Language Processing}, 17(8):1471--1482.

\bibitem[{Liu et~al.(2021)Liu, Jimerson, and Prud{'}hommeaux}]{liu-etal-2021-morphological}
Zoey Liu, Robert Jimerson, and Emily Prud{'}hommeaux. 2021.
\newblock \href {https://doi.org/10.18653/v1/2021.americasnlp-1.10} {Morphological segmentation for {S}eneca}.
\newblock In \emph{Proceedings of the First Workshop on Natural Language Processing for Indigenous Languages of the Americas}, pages 90--101, Online. Association for Computational Linguistics.

\bibitem[{Liu et~al.(2023)Liu, Spence, and Prud{'}hommeaux}]{liu-etal-2023-investigating}
Zoey Liu, Justin Spence, and Emily Prud{'}hommeaux. 2023.
\newblock \href {https://aclanthology.org/2023.eacl-main.10} {Investigating data partitioning strategies for crosslinguistic low-resource {ASR} evaluation}.
\newblock In \emph{Proceedings of the 17th Conference of the European Chapter of the Association for Computational Linguistics}, pages 123--131, Dubrovnik, Croatia. Association for Computational Linguistics.

\bibitem[{Mager et~al.(2020)Mager, {\c{C}}etino{\u{g}}lu, and Kann}]{mager-etal-2020-tackling}
Manuel Mager, {\"O}zlem {\c{C}}etino{\u{g}}lu, and Katharina Kann. 2020.
\newblock \href {https://doi.org/10.18653/v1/2020.emnlp-main.423} {Tackling the low-resource challenge for canonical segmentation}.
\newblock In \emph{Proceedings of the 2020 Conference on Empirical Methods in Natural Language Processing (EMNLP)}, pages 5237--5250, Online. Association for Computational Linguistics.

\bibitem[{Mager et~al.(2022)Mager, Oncevay, Mager, Kann, and Vu}]{mager-etal-2022-bpe}
Manuel Mager, Arturo Oncevay, Elisabeth Mager, Katharina Kann, and Thang Vu. 2022.
\newblock \href {https://doi.org/10.18653/v1/2022.findings-acl.78} {{BPE} vs. morphological segmentation: A case study on machine translation of four polysynthetic languages}.
\newblock In \emph{Findings of the Association for Computational Linguistics: ACL 2022}, pages 961--971, Dublin, Ireland. Association for Computational Linguistics.

\bibitem[{Marcus et~al.(1993)Marcus, Santorini, and Marcinkiewicz}]{marcus-etal-1993-building}
Mitchell~P. Marcus, Beatrice Santorini, and Mary~Ann Marcinkiewicz. 1993.
\newblock \href {https://aclanthology.org/J93-2004} {Building a large annotated corpus of {E}nglish: The {P}enn {T}reebank}.
\newblock \emph{Computational Linguistics}, 19(2):313--330.

\bibitem[{Mott et~al.(2020)Mott, Bies, Strassel, Kodner, Richter, Xu, and Marcus}]{mott-etal-2020-morphological}
Justin Mott, Ann Bies, Stephanie Strassel, Jordan Kodner, Caitlin Richter, Hongzhi Xu, and Mitchell Marcus. 2020.
\newblock \href {https://aclanthology.org/2020.lrec-1.493} {Morphological segmentation for low resource languages}.
\newblock In \emph{Proceedings of the Twelfth Language Resources and Evaluation Conference}, pages 3996--4002, Marseille, France. European Language Resources Association.

\bibitem[{Ott et~al.(2019)Ott, Edunov, Baevski, Fan, Gross, Ng, Grangier, and Auli}]{ott-etal-2019-fairseq}
Myle Ott, Sergey Edunov, Alexei Baevski, Angela Fan, Sam Gross, Nathan Ng, David Grangier, and Michael Auli. 2019.
\newblock \href {https://doi.org/10.18653/v1/N19-4009} {fairseq: A fast, extensible toolkit for sequence modeling}.
\newblock In \emph{Proceedings of the 2019 Conference of the North {A}merican Chapter of the Association for Computational Linguistics (Demonstrations)}, pages 48--53, Minneapolis, Minnesota. Association for Computational Linguistics.

\bibitem[{Rajpurkar et~al.(2016)Rajpurkar, Zhang, Lopyrev, and Liang}]{rajpurkar-etal-2016-squad}
Pranav Rajpurkar, Jian Zhang, Konstantin Lopyrev, and Percy Liang. 2016.
\newblock \href {https://doi.org/10.18653/v1/D16-1264} {{SQ}u{AD}: 100,000+ questions for machine comprehension of text}.
\newblock In \emph{Proceedings of the 2016 Conference on Empirical Methods in Natural Language Processing}, pages 2383--2392, Austin, Texas. Association for Computational Linguistics.

\bibitem[{S{\o}gaard et~al.(2021)S{\o}gaard, Ebert, Bastings, and Filippova}]{sogaard-etal-2021-need}
Anders S{\o}gaard, Sebastian Ebert, Jasmijn Bastings, and Katja Filippova. 2021.
\newblock \href {https://doi.org/10.18653/v1/2021.eacl-main.156} {We need to talk about random splits}.
\newblock In \emph{Proceedings of the 16th Conference of the European Chapter of the Association for Computational Linguistics: Main Volume}, pages 1823--1832, Online. Association for Computational Linguistics.

\bibitem[{Spiegler et~al.(2010)Spiegler, van~der Spuy, and Flach}]{spiegler-etal-2010-ukwabelana}
Sebastian Spiegler, Andrew van~der Spuy, and Peter~A. Flach. 2010.
\newblock \href {https://aclanthology.org/C10-1115} {{U}kwabelana - an open-source morphological {Z}ulu corpus}.
\newblock In \emph{Proceedings of the 23rd International Conference on Computational Linguistics (Coling 2010)}, pages 1020--1028, Beijing, China. Coling 2010 Organizing Committee.

\bibitem[{Vasquez et~al.(2018)Vasquez, Ego~Aguirre, Angulo, Miller, Villanueva, Agi{\'c}, Zariquiey, and Oncevay}]{vasquez-etal-2018-toward}
Alonso Vasquez, Renzo Ego~Aguirre, Candy Angulo, John Miller, Claudia Villanueva, {\v{Z}}eljko Agi{\'c}, Roberto Zariquiey, and Arturo Oncevay. 2018.
\newblock \href {https://doi.org/10.18653/v1/W18-6018} {Toward {U}niversal {D}ependencies for {S}hipibo-{K}onibo}.
\newblock In \emph{Proceedings of the Second Workshop on Universal Dependencies ({UDW} 2018)}, pages 151--161, Brussels, Belgium. Association for Computational Linguistics.

\bibitem[{Wang et~al.(2018)Wang, Singh, Michael, Hill, Levy, and Bowman}]{wang-etal-2018-glue}
Alex Wang, Amanpreet Singh, Julian Michael, Felix Hill, Omer Levy, and Samuel Bowman. 2018.
\newblock \href {https://doi.org/10.18653/v1/W18-5446} {{GLUE}: A multi-task benchmark and analysis platform for natural language understanding}.
\newblock In \emph{Proceedings of the 2018 {EMNLP} Workshop {B}lackbox{NLP}: Analyzing and Interpreting Neural Networks for {NLP}}, pages 353--355, Brussels, Belgium. Association for Computational Linguistics.

\bibitem[{Warstadt et~al.(2020)Warstadt, Parrish, Liu, Mohananey, Peng, Wang, and Bowman}]{warstadt-etal-2020-blimp-benchmark}
Alex Warstadt, Alicia Parrish, Haokun Liu, Anhad Mohananey, Wei Peng, Sheng-Fu Wang, and Samuel~R. Bowman. 2020.
\newblock \href {https://doi.org/10.1162/tacl_a_00321} {{BL}i{MP}: The benchmark of linguistic minimal pairs for {E}nglish}.
\newblock \emph{Transactions of the Association for Computational Linguistics}, 8:377--392.

\bibitem[{Woodward(1953)}]{woodward1953}
Mary~F. Woodward. 1953.
\newblock \emph{Survey of California and Other Indian Languages}.
\newblock University of California Berkeley, Woodward.002.

\end{thebibliography}

\appendix

\section{Notes on computing time and infrastructure}
\label{sec:computing}

All experiments are run on a research computing cluster. The models applied here are open-source (Section~\ref{architectures}). Given a new test sample size and a data partitioning strategy, with \textsc{CRF}, the total computing time for training the models from differently sized training sets as well as generating predictions for the evaluation sets ranges from less than a minute for Hupa, to around 20m for Zulu; with each of the seq2seq models, the total computing time spans from 6h30m for Hupa, to one day and a half for Zulu. All models are trained in sequential order with a single GPU with 8GB of memory.

\section{Detailed Results}
\label{sec:appendix_detailed} 
%\bonnieshort{Oddly in a COLING workshop submission last month we got dinged because we included an appendix; they said we were only allowed to do this for the final/camera-ready version. (Really odd.) Double check that NAACL allows this; I'm pretty sure they do, but just in case, it's worth another check on it.}

This appendix contains Table~\ref{detail_random}-\ref{detail_adversarial}.

\section{Regression results}
\label{sec:regression_full}

Regression coefficients of other control variables are presented in Table~\ref{regression_full}: (1) the method for deriving new test samples; (2) morpheme overlap; (3) the relative ratios between the training and the eval sets for the number of words; (4) the average number of morphemes per word; (5) and the average number of morpheme types per word.

\begin{table*}[h!]
\footnotesize
\centering
\begin{tabular}{lllllll}
\hline
\textbf{Language} &  \textbf{Data partitioning strategy} & \textbf{\textsc{CRF}} & \textbf{\textsc{LSTM}} & \textbf{\textsc{Transformer\_Tiny}} & \textbf{\textsc{Transformer}}  \\\hline  
Mexicanero & random  &   0.85       (\textit{0.02})   &    0.69       (\textit{0.05})    &   0.75       (\textit{0.04})    &   0.58       (\textit{0.04})      \\
& adversarial &0.85        (\textit{0.02})     &   0.67        (\textit{0.06})     &   0.74        (\textit{0.04})     &   0.57        (\textit{0.04})      \\ 
Nahuatl & random  &0.75    (\textit{0.03})  &  0.62    (\textit{0.04})  &  0.65    (\textit{0.04}) &  0.51    (\textit{0.04})   \\
&   adversarial &0.74        (\textit{0.03})      &  0.6         (\textit{0.04})    &    0.63        (\textit{0.04})   &     0.49        (\textit{0.04})      \\
Yorem Nokki &   random & 0.80    (\textit{0.02})  &  0.70    (\textit{0.04}) &  0.73   (\textit{0.03})  & 0.61   (\textit{0.04})  \\
&     adversarial & 0.79        (\textit{0.02})   &     0.68        (\textit{0.04})  &     0.72        (\textit{0.03})  &      0.59        (\textit{0.04})   \\  
Wixarika & random  & 0.76     (\textit{0.02}) &    0.7      (\textit{0.03}) &    0.69     (\textit{0.03})    & 0.59     (\textit{0.03})    \\
&    adversarial &0.76        (\textit{0.02})       & 0.68        (\textit{0.03})    &    0.68        (\textit{0.03})  &      0.58       (\textit{0.03})  \\     
Raramuri &   random &0.75   (\textit{0.03})  & 0.63   (\textit{0.05}) &  0.69   (\textit{0.03})  & 0.52   (\textit{0.08})  \\
&     adversarial & 0.74        (\textit{0.03})   &     0.61        (\textit{0.05})     &   0.67        (\textit{0.03})  &      0.50         (\textit{0.07})  \\     
Popoluca & random &  0.88     (\textit{0.02})   &  0.55     (\textit{0.05})   &  0.60      (\textit{0.04})  &   0.50      (\textit{0.04})    \\
&  adversarial &0.88        (\textit{0.02})      &  0.54        (\textit{0.05})       & 0.58        (\textit{0.03})     &   0.49        (\textit{0.03})   \\   
Tepehua & random  &0.79    (\textit{0.03})  &  0.50     (\textit{0.04})  &  0.49    (\textit{0.04}) &   0.40     (\textit{0.03}) \\  
  &   adversarial& 0.78        (\textit{0.03})    &    0.49        (\textit{0.04})    &    0.46        (\textit{0.09})    &    0.39        (\textit{0.04}) \\  
Shipibo-Konibo &    random & 0.85   (\textit{0.02}) & 0.50    (\textit{0.04})   &0.53   (\textit{0.04}) &  0.42   (\textit{0.03})  \\
&       adversarial &0.84        (\textit{0.02})     &   0.48        (\textit{0.04})     &   0.52        (\textit{0.04})     &   0.41        (\textit{0.03})       \\
Seneca & random &0.95   (\textit{0.01})  & 0.92   (\textit{0.03})  & 0.91   (\textit{0.03})  & 0.78   (\textit{0.01})  \\
&    adversarial& 0.95        (\textit{0.01})   &     0.92        (\textit{0.01})    &    0.90         (\textit{0.02})     &   0.76        (\textit{0.01})  \\     
Hupa &  random &0.78   (\textit{0.03})   &0.57   (\textit{0.05})   &0.60    (\textit{0.04})  & 0.46   (\textit{0.06})  \\
&      adversarial& 0.77        (\textit{0.03})     &   0.56        (\textit{0.05})    &    0.58        (\textit{0.04})       & 0.45        (\textit{0.06})    \\   
English & random  &0.76    (\textit{0.02})  &  0.65    (\textit{0.05})  &  0.66    (\textit{0.10}) &    0.53    (\textit{0.04})   \\
&  adversarial& 0.75        (\textit{0.02})    &    0.64        (\textit{0.05}) &       0.67        (\textit{0.05}) &       0.52        (\textit{0.04})   \\        
German & random &0.73   (\textit{0.02}) &  0.66   (\textit{0.04}) &  0.64   (\textit{0.06}) &  0.53   (\textit{0.04})  \\
&     adversarial &0.72        (\textit{0.02})    &   0.64        (\textit{0.04})   &     0.63        (\textit{0.04})   &     0.53        (\textit{0.03})    \\    
Finnish & random  &0.74    (\textit{0.03})  &  0.60     (\textit{0.04})   & 0.59    (\textit{0.04})  &  0.46    (\textit{0.03})   \\
&   adversarial &0.73        (\textit{0.03})      &  0.59        (\textit{0.05}) &       0.58        (\textit{0.05})    &    0.45        (\textit{0.03})   \\
Turkish & random & 0.74    (\textit{0.02}) &   0.65    (\textit{0.04})  &  0.68    (\textit{0.07})   & 0.58    (\textit{0.03})   \\
&   adversarial &0.73        (\textit{0.02})    &    0.64        (\textit{0.03})    &    0.68        (\textit{0.02})   &     0.56        (\textit{0.02})    \\ 
Indonesian & random  &   0.90        (\textit{0.01})    &   0.86       (\textit{0.02})    &   0.88       (\textit{0.05})   &    0.68       (\textit{0.03})     \\ 
& adversarial& 0.89        (\textit{0.01})    &    0.86        (\textit{0.02})   &    0.87        (\textit{0.04})     &   0.66        (\textit{0.03})   \\ 
Zulu  & random & 0.85   (\textit{0.01})  & 0.84   (\textit{0.01})  & 0.79   (\textit{0.01})  & 0.64   (\textit{0.05})  \\
&     adversarial & 0.85        (\textit{0.01})     &   0.83        (\textit{0.01})   &     0.78        (\textit{0.01})   &     0.63        (\textit{0.05})    \\
Akan &   random & 0.82   (\textit{0.01}) &  0.76   (\textit{0.02}) &  0.67   (\textit{0.15}) &  0.70    (\textit{0.02})  \\
&         adversarial & 0.81        (\textit{0.01})   &     0.75        (\textit{0.02})    &    0.69        (\textit{0.14})   &     0.70         (\textit{0.02})  \\     
Swahili &   random & 0.77   (\textit{0.02})  & 0.68   (\textit{0.10}) &   0.68   (\textit{0.11})  & 0.57   (\textit{0.12})  \\
&       adversarial & 0.76        (\textit{0.02}) &       0.68        (\textit{0.09})    &    0.66        (\textit{0.12})  &      0.57        (\textit{0.11})   \\    
Telugu &  random & 0.71   (\textit{0.02})  & 0.64   (\textit{0.04})  & 0.68   (\textit{0.03}) &  0.53   (\textit{0.04})  \\
&       adversarial & 0.69        (\textit{0.02})     &   0.59       (\textit{0.11})    &    0.64        (\textit{0.09})   &     0.46        (\textit{0.10})        \\
\hline
\end{tabular}
\caption{
Language-by-language $F1$ scores averaged across differently sized new test samples, given each combination of a data partitioning strategy, test set proportion, and model architecture. Here new test samples are generated \textbf{randomly}. Standard deviations are (\textit{italicized}).}
\label{detail_random}
\end{table*}

\newpage

\begin{table*}[h!]
\footnotesize
\centering
\begin{tabular}{lllllll}
\hline
\textbf{Language} &  \textbf{Data partitioning strategy} & \textbf{\textsc{CRF}} & \textbf{\textsc{LSTM}} & \textbf{\textsc{Transformer\_Tiny}} & \textbf{\textsc{Transformer}}  \\\hline      
Mexicanero & random  &   0.55       (\textit{0.20})   &     0.51       (\textit{0.16})   &    0.5        (\textit{0.16})   &    0.41       (\textit{0.14})     \\ 
& adversarial &0.46        (\textit{0.21})      &  0.43        (\textit{0.18})      &  0.42        (\textit{0.18})   &     0.36        (\textit{0.15})     \\  
Nahuatl & random  &0.52    (\textit{0.14}) &   0.44    (\textit{0.12})  &  0.44    (\textit{0.09})  &  0.35    (\textit{0.07})   \\
&   adversarial& 0.45        (\textit{0.15})    &    0.37       (\textit{0.11})     &   0.38        (\textit{0.08})  &      0.32        (\textit{0.07})       \\
Yorem Nokki &  random &0.53   (\textit{0.16}) &  0.42   (\textit{0.12})  & 0.44   (\textit{0.12}) &   0.34   (\textit{0.10})   \\
&       adversarial &0.46        (\textit{0.14})   &     0.38        (\textit{0.11})    &    0.42        (\textit{0.10})   &      0.32        (\textit{0.07})    \\  
Wixarika & random &  0.68     (\textit{0.07}) &     0.61     (\textit{0.10})   &   0.58     (\textit{0.08})  &  0.50      (\textit{0.06})    \\
&  adversarial &0.64        (\textit{0.07})   &    0.56        (\textit{0.08})   &     0.52        (\textit{0.08})   &     0.46        (\textit{0.03})    \\  
Raramuri &  random &0.52   (\textit{0.12})  & 0.4    (\textit{0.12})  & 0.45   (\textit{0.12})  & 0.36   (\textit{0.09})  \\
&     adversarial & 0.46       (\textit{ 0.10})  &       0.33        (\textit{0.06})     &   0.39        (\textit{0.08})     &   0.31        (\textit{0.07})     \\  
Popoluca & random &  0.80      (\textit{0.05})  &   0.39     (\textit{0.11})  &   0.42     (\textit{0.09})   &  0.36     (\textit{0.10})     \\
&  adversarial &0.78        (\textit{0.07})      &  0.35        (\textit{0.09})    &    0.40         (\textit{0.07})   &     0.34        (\textit{0.08})    \\  
Tepehua & random & 0.68    (\textit{0.10})  &   0.39    (\textit{0.09})   & 0.39    (\textit{0.07}) &   0.37    (\textit{0.07})   \\
&     adversarial& 0.63        (\textit{0.11})    &    0.34        (\textit{0.07})     &   0.35        (\textit{0.06})    &    0.32        (\textit{0.05})    \\ 
Shipibo-Konibo &   random& 0.64   (\textit{0.10})  &  0.32   (\textit{0.08}) &  0.33   (\textit{0.08})  & 0.27   (\textit{0.06})  \\
&        adversarial &0.53        (\textit{0.17})   &     0.25       (\textit{0.08})     &   0.26        (\textit{0.09})    &    0.21        (\textit{0.07})    \\   
Seneca & random  & 0.89   (\textit{0.05}) &  0.82   (\textit{0.09}) &  0.79   (\textit{0.07})  & 0.59   (\textit{0.05})  \\
&  adversarial &0.86        (\textit{0.04})    &    0.78       (\textit{0.08})    &    0.75        (\textit{0.05})       & 0.56        (\textit{0.04})  \\     
Hupa  & random &0.67   (\textit{0.11})  & 0.50    (\textit{0.14}) &  0.48   (\textit{0.11}) &  0.39   (\textit{0.11})  \\
&     adversarial &0.63        (\textit{0.10})        & 0.45        (\textit{0.11})       & 0.45        (\textit{0.08})  &      0.36        (\textit{0.10})  \\      
English & random & 0.52    (\textit{0.05})  &  0.41    (\textit{0.05})  &  0.43    (\textit{0.04})   & 0.31    (\textit{0.04})   \\
&    adversarial &0.48        (\textit{0.08})     &   0.35        (\textit{0.08})      &  0.37        (\textit{0.07})    &    0.27        (\textit{0.05})  \\        
German & random &0.63   (\textit{0.10}) &   0.54   (\textit{0.10})  & 0.52   (\textit{0.09}) &  0.43   (\textit{0.08})  \\
&    adversarial &0.59        (\textit{0.09})     &   0.49        (\textit{0.07})     &   0.47        (\textit{0.08})       & 0.38        (\textit{0.07})      \\ 
Finnish & random  &0.63    (\textit{0.07})  &  0.49    (\textit{0.07})  &  0.45    (\textit{0.05})   & 0.34    (\textit{0.04})   \\
&  adversarial & 0.60         (\textit{0.08})    &    0.46        (\textit{0.07}) &       0.40         (\textit{0.06})    &    0.31        (\textit{0.04})  \\  
Turkish & random  &0.73    (\textit{0.07}) &   0.57    (\textit{0.09}) &   0.58    (\textit{0.06})  &  0.46    (\textit{0.06})   \\
&   adversarial& 0.69        (\textit{0.09})  &      0.52        (\textit{0.09})   &     0.53        (\textit{0.06}) &       0.43        (\textit{0.07})    \\   
Indonesian & random   &  0.60        (\textit{0.19})   &    0.53       (\textit{0.21})    &   0.56       (\textit{0.19})      & 0.39       (\textit{0.20}) \\      
&  adversarial& 0.42        (\textit{0.24})     &   0.36        (\textit{0.24})   &     0.37        (\textit{0.25}) &       0.23        (\textit{0.12})    \\   
Zulu & random& 0.78   (\textit{0.06})  & 0.73   (\textit{0.07})  & 0.69   (\textit{0.05}) &  0.49   (\textit{0.04})  \\
&     adversarial& 0.74        (\textit{0.06})    &    0.68        (\textit{0.08})     &   0.68        (\textit{0.05})  &      0.49        (\textit{0.05})  \\
Akan &   random &0.77   (\textit{0.14})  & 0.67  (\textit{0.13}) &  0.67   (\textit{0.14})  & 0.61   (\textit{0.10})   \\
&     adversarial& 0.68        (\textit{0.22})    &    0.58        (\textit{0.16})     &   0.62        (\textit{0.15})   &     0.55        (\textit{0.12})      \\ 
Swahili &   random &0.71   (\textit{0.11}) &  0.66   (\textit{0.11}) &  0.67   (\textit{0.11}) &  0.58   (\textit{0.11})  \\
&       adversarial& 0.65        (\textit{0.09})    &    0.63        (\textit{0.10})   &      0.62        (\textit{0.11})    &    0.55        (\textit{0.10})    \\    
Telugu &   random &0.44   (\textit{0.08}) &  0.40    (\textit{0.10}) &   0.43   (\textit{0.09})  & 0.33   (\textit{0.09})  \\
&      adversarial& 0.37        (\textit{0.07})   &     0.32        (\textit{0.06})   &     0.35       (\textit{0.06})     &  0.26        (\textit{0.06})  \\
\hline
\end{tabular}
\caption{
Language-by-language $F1$ scores averaged across differently sized new test samples, given each combination of a data partitioning strategy, test set proportion, and model architecture. Here new test samples are generated \textbf{adversarially}. Standard deviations are (\textit{italicized}).}
\label{detail_adversarial}
\end{table*}

\clearpage

\begin{table*}[h!]
\footnotesize
%\resizebox{\columnwidth}{!}{
\centering
\begin{tabular}{lllllll}
\hline
\textbf{Language}  & \textbf{Randomly generating} & \textbf{Morph} & \textbf{$N$ of words} & \textbf{Avg. $N$ of morph} & \textbf{Avg. $N$ of morph type} & $R^2$\\
&  \textbf{new test samples} & \textbf{overlap} & &  \textbf{per word} & \textbf{per word} & \\
\hline
Mexicanero & 0.51*** & 0.56*** & 1.42*** & 3.05*** & -4.19*** & 0.87 \\
Nahuatl &  0.69*** & 0.98*** & 0.58*** & 2.45*** & -2.09*** & 0.90 \\
Yorem Nokki & 0.52*** & 0.87*** & -0.54*** & 0.93*** & 0.54** & 0.82 \\
Wixarika & -0.16*** & -0.40*** & -0.72*** & -2.74*** & 2.99*** & 0.72 \\
Raramuri & 0.59*** & 1.22*** & 0.79*** & 2.57*** & -2.78*** & 0.87 \\
Popoluca & 0.57*** & 0.68*** &  -0.29*** & 0.03 & 0.33* & 0.94 \\
Tepehua & 0.41*** & 0.66*** & -0.56*** & -0.72*** & 1.30*** & 0.92 \\
Shipibo-Konibo & 0.98*** & 1.29*** & 0.83*** & 2.91*** & -2.64*** & 0.95 \\
Seneca & 3.66*** & 1.29*** & 3.33*** & 4.27*** & -4.48*** \\
Hupa & 0.47*** & 0.95*** & 0.33*** & 1.53*** & -1.72*** & 0.92 \\
English & 0.86*** & 1.17*** & 0.66*** & 2.00*** & -1.72*** & 0.92 \\
German & -0.02 & 0.72*** & -0.84*** & -0.54*** & 1.57*** & 0.89 \\
Finnish & 0.65*** & 0.79*** & 0.94*** & 2.00*** & -2.03*** & 0.93 \\
Turkish & 0.85*** & 0.71*** & 1.25*** & 2.62*** & -3.02*** & 0.86 \\
Indonesian & 1.18*** & 1.14*** & 0.09 & 2.20*** & -0.51* & 0.93 \\
Zulu & 1.96*** & 1.16*** & 2.10*** & 3.98*** & -4.22*** & 0.90 \\
Akan & 1.92*** & 1.43*** & 1.90*** & 3.61*** & -4.11*** & 0.59 \\
Swahili & 0.60*** & 0.89*** & 0.59* & 1.89*** & -1.78** & 0.51 \\
Telugu & 0.58*** & 0.64*** & 0.14 & 1.43*** & -0.77** & 0.84 \\
\hline
\end{tabular}
%}
\caption{
Regression coefficients of other control variables and $R^2$ of the regression model for each language; the number of * suggests significance level: * $p < 0.05$, ** $p < 0.01$, *** $p < 0.001$.}
\label{regression_full}
\end{table*}

\end{document}